\documentclass{article}

\PassOptionsToPackage{numbers, compress}{natbib}
\usepackage{wrapfig}
\usepackage[preprint]{neurips_2025}

\usepackage{caption}

\usepackage[utf8]{inputenc} % allow utf-8 input
\usepackage[T1]{fontenc}    % use 8-bit T1 fonts
\usepackage{hyperref}       % hyperlinks
\usepackage{url}            % simple URL typesetting
\usepackage{booktabs}       % professional-quality tables
\usepackage{amsfonts}       % blackboard math symbols
\usepackage{nicefrac}       % compact symbols for 1/2, etc.
\usepackage{microtype}      % microtypography
\usepackage{xcolor}         % colors

\usepackage{graphicx} 
\usepackage{hyperref}
\usepackage{multirow}
\usepackage{amsmath}
\usepackage{amssymb}

\newcommand{\ie}{\textit{i.e.}}
\newcommand{\etal}{\textit{et. al}}

\newcommand{\aka}{\textit{a.k.a.}}

\title{Learning Robust Anymodal Segmentor with Unimodal and Cross-modal Distillation}
\author{
Xu Zheng$^{1}$ \quad
Haiwei Xue$^{2,1}$ \quad
Jialei Chen$^{3}$ \quad
Yibo Yan$^{1}$ \quad
Lutao Jiang$^{1}$ \quad
Yuanhuiyi Lyu$^{1}$ \quad \\ 
\textbf{Kailun Yang}$^{4}$ \quad 
\textbf{Linfeng Zhang}$^{5}$ \quad
\textbf{Xuming Hu}$^{1}$\thanks{Corresponding author: xuminghu@hkust-gz.edu.cn} \\ \\
$^{1}$HKUST(GZ) \quad
$^{2}$Tsinghua University \quad
$^{3}$Nagoya University \\
$^{4}$Hunan University \quad
$^{5}$Shanghai Jiao Tong University
}

\begin{document}
\maketitle
\begin{abstract}
% Simultaneously using multimodal inputs from multiple sensors to train segmentors is intuitively advantageous but practically challenging. A key challenge is unimodal bias, where multimodal segmentors over-rely on certain modalities, causing performance drops when these modalities are missing—common in real-world applications. To this end, we develop the \textbf{first} framework for learning robust segmentor that can handling any combinations of visual modalities. Specifically, we first introduce a parallel multimodal learning strategy for learning a strong teacher. The cross-modal and unimodal distillation is then achieved in the multiscale representation space by transferring the feature-level knowledge from multimodal to anymodal segmentors, aiming at addressing the unimodal bias and avoiding over reliance on specific modalities. Moreover, a prediction-level modality-agnostic semantic distillation is proposed to achieve semantic knowledge transferring for segmentation. Extensive experiments on both synthetic and real world multi-sensor benchmarks demonstrate that our \textbf{\textit{AnySeg}} achieves superior performance (\textbf{+6.37\% \& +6.15\%}).
Leveraging multimodal inputs from multiple sensors offers intuitive benefits for semantic segmentation but introduces practical challenges—most notably, unimodal bias, where models overfit to dominant modalities and perform poorly when others are missing, a common issue in real-world scenarios. To address this, we propose \textit{AnySeg}, a unified framework for learning robust segmentors that generalize to arbitrary combinations of input modalities. Our approach first trains a strong multimodal teacher using parallel modality learning. We then distill both unimodal and cross-modal knowledge to an anymodal student via multiscale feature-level distillation, reducing modality dependence and improving generalization. To further enhance semantic consistency, we introduce a prediction-level, modality-agnostic distillation loss. Unlike prior work, our framework explicitly handles missing modalities challenges by learning \textbf{\textit{unimodal and cross-modal}} correspondence among input modalities. Extensive experiments on synthetic and real-world multi-sensor datasets demonstrate the effectiveness of AnySeg, achieving notable improvements of \textbf{+6.37\%} and \textbf{+6.15\%} in mIoU.

\end{abstract}

\section{Introduction}
The success of multimodal deep learning relies heavily on effectively leveraging information from multiple modalities, particularly for complex tasks such as semantic segmentation in scene understanding~\cite{zheng2024360sfuda++,lyu2024omnibind,zheng2024transformer,lyu2024unibind}. While intuitively beneficial, training segmentation models, \aka, segmentors, with inputs from multiple sensors presents significant practical challenges. A prominent issue in this context is \textbf{\textit{unimodal bias}}~—~a phenomenon where networks develop an over-reliance on a single, faster-to-learn modality, often overlooking other sources of valuable information. This bias stems from the distinct characteristics and varied learning dynamics of each sensor modality.

For example, the well-known CMX model~\cite{zhang2023cmx} in multimodal semantic segmentation suffers significant performance drops when evaluated without the RGB modality. Meanwhile, the state-of-the-art model Any2Seg~\cite{zheng2024learning}, which aims at learning modality-agnostic representation for missing modality problems, demonstrates a significant performance decline when evaluated in modality-incomplete scenarios. For instance, when depth data is missing, segmentation performance drops markedly (RD: 68.21 $\rightarrow$ R: 39.02, a \textit{\textbf{decrease}} of \textbf{\textit{29.19}} mIoU), illustrating how unimodal bias can lead to substantial performance degradation in real-world applications where certain modalities are often unavailable.
% \begin{figure}[t!]
%     \centering
%     \includegraphics[width=0.5\linewidth]{images/teaser.pdf}
%     % \vspace{-24pt}
%     \caption{Comparison between ours against SoTA methods~\cite{zheng2024centering,zheng2024learning} on (a) MUSES and (b) DELIVER datasets.
%     }
%     % \vspace{-20pt}
%     \label{fig:teaser}
% \end{figure}

Despite advancements in multimodal learning, such as leveraging large multimodal language models~\cite{zheng2024learning} and prioritizing each modality~\cite{zheng2024centering}, progress in addressing unimodal bias and fostering robust multimodal correlations remains limited. To address this gap, we introduce the first framework for learning robust anymodal segmentors\footnote{We define anymodal segmentors as models that ensure robust performance despite missing modalities.}. This framework is tailored to handle real-world scenarios where modality completeness cannot be guaranteed, such as missing modality~\cite{liu2024fourier} or modality-agnostic segmentation~\cite{zheng2024learning}.

Our approach begins with a novel \textbf{P}arallel \textbf{M}ultimodal \textbf{L}earning (\textbf{PML}) strategy, which facilitates the learning of a strong teacher model for both unimodal and multimodal distillation \textit{without adding extra parameters}. Inspired by recent methods~\cite{zheng2024learning, zheng2024centering}, we process all multimodal inputs from different sensors in a single mini-batch, passing them through the segmentation backbone, \ie, SegFormer. Multimodal fusion is performed through simple averaging, and supervision is applied at the final layer of the segmentation decoder. This straightforward yet effective PML strategy enables segmentor to focus on capturing both unimodal and multimodal knowledge (See Tab.~\ref{tab:ablation_loss_umd} and Tab.~\ref{tab:ablation_loss_cmd_baseumd}).

We then introduce a dual-level distillation process: Unimodal Distillation (UMD) and Cross-modal Distillation (CMD), applied across multi-scale representations and prediction levels. To simulate real-world scenarios, we apply an anymodal dropout strategy, where the multimodal inputs are randomly masked, creating varied modality combinations within each batch. For distribution distillation within the multi-scale representation space, the features from the anymodal segmentor are trained to align with the corresponding features from the multimodal teacher, thereby replicating the unimodal feature extraction capabilities. Furthermore, cross-modal correspondence is applied across all active modalities to mitigate the effects of unimodal bias.
Finally, at the prediction level, we employ modality-agnostic semantic distillation to facilitate effective task-specific knowledge transfer between teacher and student models, further enhancing the robustness in diverse real-world conditions.

Extensive experiments on real-world and synthetic benchmarks demonstrate the superior robustness and performance of our method compared to existing state-of-the-art approaches, achieving mIoU improvements of \textbf{+6.37\%} and \textbf{+6.15\%}, respectively.
%\footnote{\url{https://github.com/zhengxuJosh/AnySeg}}.  
Moreover, we analyze why fused multimodal fusion distillation is \textbf{unsuitable} for ensuring robustness in multimodal segmentation and further discuss the feature characteristics of multimodal data.

\section{Related Work}
\noindent \textbf{Multimodal Semantic Segmentation}
Semantic segmentation with multi-sensor inputs enhances scene understanding by leveraging complementary information from diverse sensors, such as event cameras~\cite{zhou2024eventbind,zheng2024eventdance}, LiDAR sensors~\cite{li2023mseg3d}, and others~\cite{liao2025memorysam,liao2025benchmarking,zheng2023both,zheng2024semantics,zheng2024360sfuda++}. Recent advances in multi-sensor systems have led to the development of various approaches~\cite{zheng2025reducing,zheng2024centering,zheng2024learning,zhang2023cmx} and datasets~\cite{zhang2023delivering, brodermann2024muses} that extend from dual-modality fusion to full multimodal fusion, with the aim of achieving robust perception across diverse lighting and environmental conditions throughout the day~\cite{zhao2025unveiling,broedermann2022hrfuser,wei2023mmanet,zhang2021abmdrnet,man2023bev,wang2022multimodal,chen2021spatial,zhang2023delivering,zhang2023cmx,zhu2024customize}. For instance, MUSES~\cite{brodermann2024muses} dataset integrates data from a frame camera, LiDAR, radar, event camera, and IMU/GNSS sensors to capture driving scenes in adverse conditions with increased uncertainty.
Recently, CMNeXt~\cite{zhang2023delivering} introduced the task of fusing an arbitrary number of modalities, although this approach still relies primarily on RGB input for optimal performance. In our work, we address the challenge of \textbf{\textit{unimodal bias}} in multimodal semantic segmentation by focusing on developing a robust anymodal segmentor that can maintain performance across various input combinations, rather than solely optimizing for multimodal segmentation accuracy.

\noindent \textbf{Missing Modality Robustness}
In the multimodal learning community, several studies have sought to understand unimodal bias from both empirical~\cite{kleinman2023critical,peng2022balanced} and theoretical perspectives~\cite{huang2022modality}. As shown by Huang \etal~\cite{huang2022modality,zhangunderstanding}, while multimodal learning has the potential to surpass unimodal performance, it often falls short due to modality competition: only the subset of modalities more closely aligned with the encoder's initial parameters tends to dominate learning within the multimodal network. 
This phenomenon also occurs in multimodal semantic segmentation, as MAGIC~\cite{zheng2024centering} and Any2Seg~\cite{zheng2024learning} struggle when depth data is missing during inference. 
In this work, we focus on addressing practical challenges in multi-sensor systems that are widely applicable across industrial domains, including autonomous driving and intelligent systems. We define the unimodal bias problems in multimodal semantic segmentation and propose the anymodal semantic segmentation framework.

\noindent \textbf{Robust Multimodal Segmentors.}
In practice, sensor failures often result in incomplete multimodal data, challenging segmentation frameworks typically trained on complete modality pairs~\cite{liu2024fourier}. Recent studies aim to build models that, while trained with full modalities, remain effective when some inputs are missing~\cite{liu2024fourier,wang2023multi,maheshwari2024missing,reza2023robust,chen2023redundancy,zhao2023multi}. 
Wang \etal~\cite{wang2023learnable} proposed adaptive modality selection and knowledge distillation for cross-modal compensation. Liu \etal~\cite{liu2024fourier} extended this to modality-incomplete scene segmentation, addressing both system- and sensor-level failures. More recent methods like MAGIC~\cite{zheng2024centering} and Any2Seg~\cite{zheng2024learning} aim for modality-agnostic segmentation by extracting shared representations. However, unimodal bias remains unresolved. For instance, Any2Seg's performance drops sharply without depth input (RD: 68.21 $\rightarrow$ R: 39.02), illustrating the challenge.
To address this, we propose the \textbf{\textit{first}} framework for learning robust anymodal segmentors capable of handling missing modalities by distilling both unimodal and cross-modal knowledge. We also introduce a parallel multimodal learning strategy to build a strong teacher model, further advancing robust segmentation under incomplete inputs.

\begin{figure*}[t!]
    \centering
    \includegraphics[width=.99\textwidth]{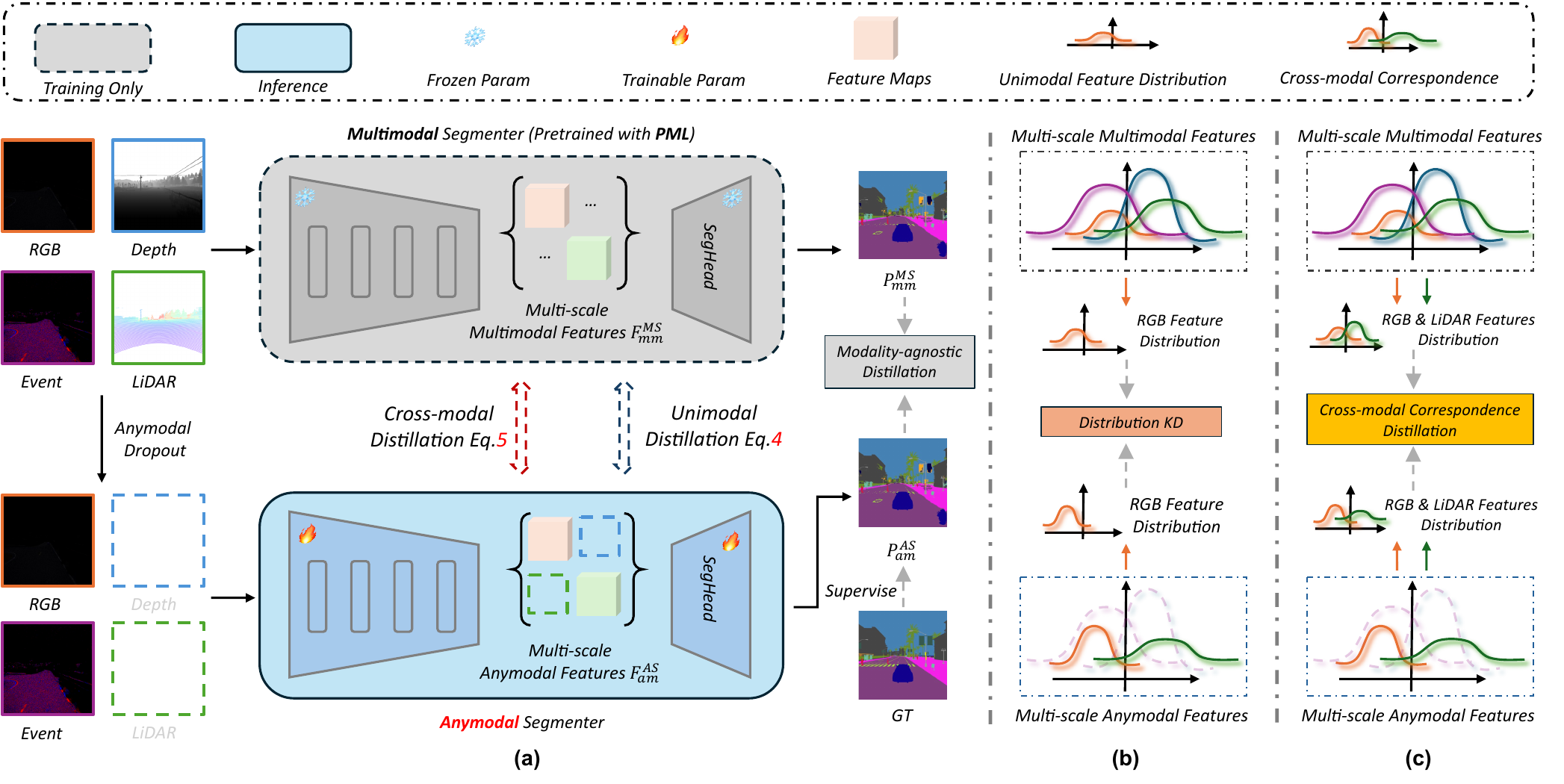}
    \vspace{-6pt}
\caption{(a) Overall of AnySeg with a two-stage training strategy: the multimodal teacher is first trained using PML, then frozen for student distillation. (b) Unimodal feature distillation transfers intra-modality knowledge. (c) Cross-modal feature distillation enables modality interaction transfer. 
}
    \vspace{-16pt}
    \label{fig:overall}
\end{figure*}
\section{Methodology}
% \subsection{Overview}

The overall framework is depicted in Fig.~\ref{fig:overall}. 
It consists of two segmentors, \ie, the multimodal teacher segmentor $\mathcal{F}_{ms}$ and anymodal student segmentor $\mathcal{F}_{as}$, as well as two key modules, including the unimodal and cross-modal distillation and the modality-agnostic semantic distillation modules. The teacher $\mathcal{F}_{ms}$ is first pre-trained with our proposed parallel multimodal learning strategy to learn a strong teacher with expertise in multimodal scenarios, its parameter is frozen during training the student $\mathcal{F}_{as}$. \noindent \textbf{Inputs:} Our framework processes multi-modal visual data from four modalities, all within the same scene. We consider RGB images \(\textbf{R} \in \mathbb{R}^{h \times w \times 3}\), depth maps \(\textbf{D} \in \mathbb{D}^{h \times w \times C^D}\), LiDAR data \(\textbf{L} \in \mathbb{L}^{h \times w \times C^P}\), and event stack images \(\textbf{E} \in \mathbb{E}^{h \times w \times C^E}\) to illustrate our method, as depicted in Fig.~\ref{fig:overall}. Here, we follow the data processing as ~\cite{zhang2023delivering}, where the channel dimensions $C^D=C^P=C^E=3$, and we also integrates the corresponding ground truth \(Y\) across \(K\) categories. For each training iteration, a mini-batch \(\{r,d,e,l\}\) contains samples from all the input modalities.

\subsection{Parallel Multimodal Learning (PML) Strategy} 
\begin{wrapfigure}{l}{8cm}
    \centering
        \centering
        \includegraphics[width=\linewidth]{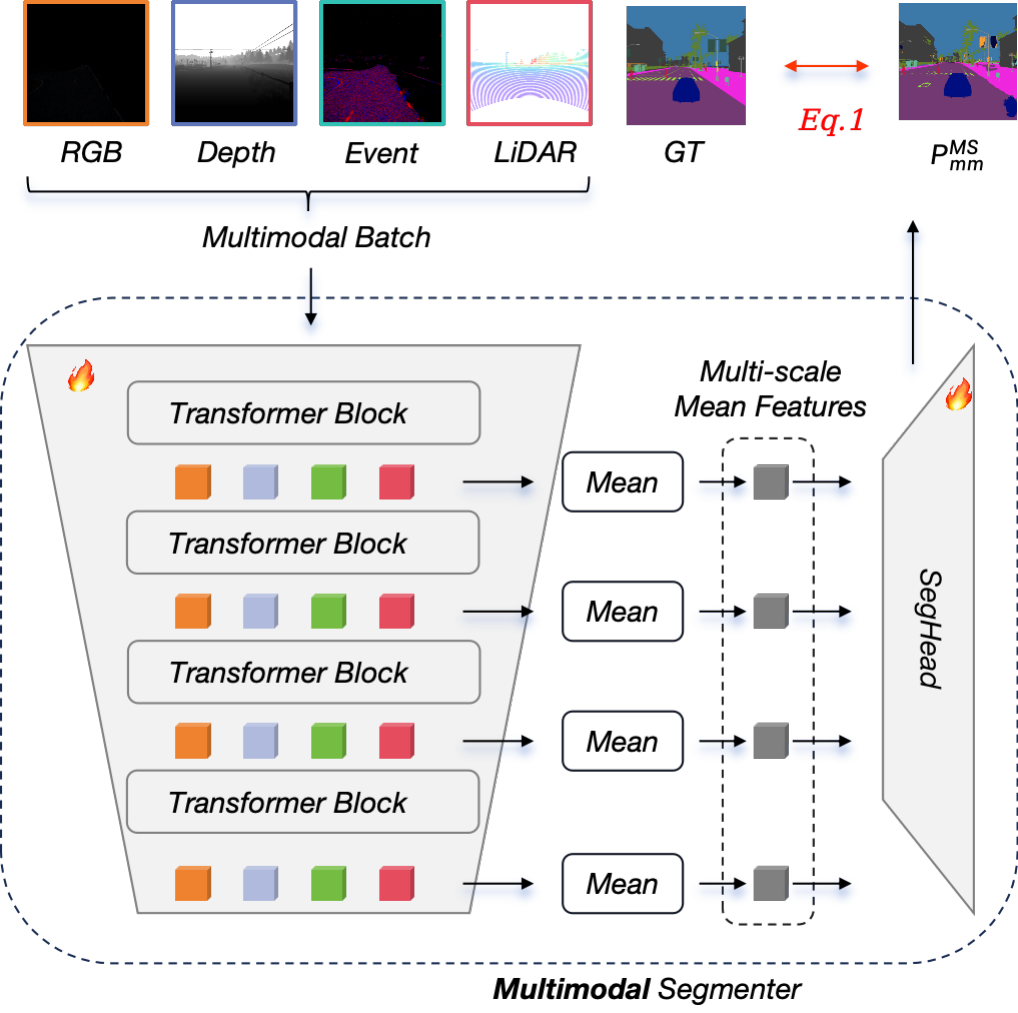}
        \caption{PML for learning a strong multimodal segmentor as teacher model.}
        \label{fig:PML}
\end{wrapfigure}

Recent studies have shown that treating all input modalities equally can enhance both multimodal and unimodal performance~\cite{zheng2024learning,zheng2024centering}. Building on insights from MAGIC~\cite{zheng2024centering}, 
we adopt a uniform approach for handling all multimodal inputs and introduce a parallel multimodal learning strategy to train a robust teacher model for knowledge distillation. 
As illustrated in Fig.~\ref{fig:PML}, we compute the mean across multimodal features at each block of the segmentation backbone~\cite{xie2021segformer}. This averaged output serves as the input for the segmentation head, leading to improved multimodal performance, particularly on real-world benchmarks, achieving 51.37 mIoU on MUSES.
The supervised loss \( L_{pre} \) for training is:
\begin{equation}
\setlength{\abovedisplayskip}{3pt}
\setlength{\belowdisplayskip}{3pt}
\mathcal{L}_{pre} = -\sum_{i=1}^{N} \sum_{k=1}^{K} y_{i,k} \log(p_{i,k}),
\end{equation}
where \( N = h \times w \) is the total number of pixels, \( y_{i,k} \) is the ground truth label for class \( k \) at pixel \( i \), and \( p_{i,k} \) is the predicted probability for class \( k \) at pixel \( i \). $\mathcal{L}_{pre}$ encourages accurate predictions across all modalities and contributes to the robustness of the teacher model.

\subsection{Unimodal and Cross-modal Distillation}
After obtaining the strong multimodal teacher $\mathcal{F}_{ms}$, we turn to learn efficient and robust anymodal segmentor $\mathcal{F}_{as}$. As depicted in Fig.~\ref{fig:overall}, the input multimodal mini-batch $\{r,d,l,e\}$ is directly ingested by weight-shared encoder within the multimodal segmentor $\mathcal{F}_{ms}$. This process yields multi-scale features $\{f_r^i,f_d^i,f_e^i,f_l^i\}^4_{i=1}$:
\begin{equation}
\{f_r^i,f_d^i,f_e^i,f_l^i\}^4_{i=1} = F_{ma}(\{r,d,l,e\}),
\end{equation}
where \( i \) represents the multi-scale feature level, summed from 1 to 4.
Meanwhile, the input multimodal mini-batch ${r,d,l,e}$ undergoes random masking to generate an anymodal batch, where modality data is randomly dropped from the batch, ensuring that at least one modality is retained in each instance. The anymodal batch is then processed by the encoder of the anymodal segmentor, yielding features ${g_r^i, g_d^i}_{i=1}^4$\footnote{For example, we illustrate the case where the event and LiDAR modalities are dropped.} for the retained modalities:
\begin{equation} 
\setlength{\abovedisplayskip}{3pt} 
\setlength{\belowdisplayskip}{3pt} \{g_r^i, g_d^i\}_{i=1}^4 = \mathcal{F}_{as}(\{r,d\}).
\end{equation}
This process ensures that the anymodal segmentor is trained on diverse input combinations, improving its robustness and adaptability to incomplete data.

\noindent \textbf{Unimodal Distillation.} After extracting the multi-scale features \( \{f_r^i, f_d^i, f_e^i, f_l^i\}_{i=1}^4 \) and \( \{g_r^i, g_d^i\}_{i=1}^4 \) from the multimodal segmentor (teacher model) and the anymodal segmentor (student model), respectively, we proceed with the distillation process. For the remaining multi-scale features \( \{g_r^i, g_d^i\}_{i=1}^4 \) from the anymodal segmentor \( \mathcal{F}_{am} \), we align them with the corresponding features \( \{f_r^i, f_d^i, f_e^i, f_l^i\}_{i=1}^4 \) obtained from the multimodal segmentor. 
% The unimodal knowledge distillation loss function based on KL divergency is then formulated as:
% \begin{equation}
% \setlength{\abovedisplayskip}{3pt}
% \setlength{\belowdisplayskip}{3pt}
% \mathcal{L}_{umd} = \sum_{i=1}^4 \left( \sum_{j=1}^{C_i} g_r^{i,j} \log\left(\frac{g_r^{i,j}}{f_r^{i,j}}\right) + \sum_{j=1}^{C_i} g_d^{i,j} \log\left(\frac{g_d^{i,j}}{f_d^{i,j}}\right) \right),
% \end{equation}
% where \( C_i \) denotes the channel count of the \( i \)-th level features.
% $\mathcal{L}_{umd}$ facilitates the transfer of knowledge within each modality, thereby enhancing the performance of the anymodal segmentor in handling single-modality data. 
The unimodal knowledge distillation loss function based on KL divergence is defined as:
\begin{equation}
\setlength{\abovedisplayskip}{3pt}
\setlength{\belowdisplayskip}{3pt}
\mathcal{L}_{umd} = \sum_{i=1}^4 \left( \sum_{j=1}^{C_i} \tilde{g}_r^{i,j} \log\left(\frac{\tilde{g}_r^{i,j}}{\tilde{f}_r^{i,j}}\right) + \sum_{j=1}^{C_i} \tilde{g}_d^{i,j} \log\left(\frac{\tilde{g}_d^{i,j}}{\tilde{f}_d^{i,j}}\right) \right),
\end{equation}
where \( C_i \) denotes the number of channels in the \( i \)-th level features. To ensure valid probability distributions for KL divergence computation, we apply \textit{softmax} to the teacher features \( g \) and \textit{log-softmax} to the student features \( f \), yielding normalized representations \( \tilde{g} \) and \( \tilde{f} \). This guarantees \textbf{\textit{non-negativity and unit sum}}, thus satisfying the theoretical prerequisites of KL divergence.
The loss term \( \mathcal{L}_{umd} \) promotes intra-modality knowledge transfer, enhancing the anymodal segmentor's ability to generalize from unimodal inputs.
This is demonstrated in Tab.~\ref{tab:ablation_loss_umd}, where we show the performance improvements achieved by applying unimodal distillation in the anymodal segmentation task. However, while unimodal knowledge transfer enhances single-modality performance, especially for the RGB images, it simultaneously hinders multimodal performance when different modality combinations are encountered, as also illustrated in Tab.~\ref{tab:ablation_loss_umd}. 

\noindent \textbf{Cross-modal Correspondence Distillation.}

While unimodal knowledge distillation significantly improves segmentation performance on RGB images, as shown in Tab.~\ref{tab:ablation_loss_umd}, it also introduces unimodal bias that reduces performance on other modalities. This bias causes the model to over-rely on the RGB modality, which is easier for the model to learn, thereby limiting its generalization capacity across diverse input types. Consequently, unimodal knowledge transfer, while beneficial for single-modality performance, negatively impacts multimodal performance in mixed-modality scenarios. This is evident in Tab.~\ref{tab:ablation_loss_umd}, where performance decreases for modalities such as Event (-3.74\% $\downarrow$) and LiDAR (-2.91\% $\downarrow$).

To address this, we leverage cross-modal correspondences between ``easy-to-learn'' and ``hard-to-learn'' modalities to achieve a more balanced performance. By distilling these cross-modal relationships from the teacher to the student model—referred to as the ``anymodal'' segmentor—we effectively mitigate unimodal bias across all modalities.
% The distillation of cross-modal knowledge between student features \(\{g_r^i, g_d^i\}_{i=1}^4\) and teacher features \(\{f_r^i, f_d^i, f_e^i, f_l^i\}_{i=1}^4\) is achieved through:
% % \begin{equation} 
% % \mathcal{L}_{cmd} = KL\left(\mathcal{S}\left(\{g_d^i\}_{i=1}^4, \{g_r^i\}_{i=1}^4\right), \, \mathcal{S}\left(\{f_d^i\}_{i=1}^4, \{f_r^i\}_{i=1}^4\right)\right),
% % \end{equation}
% \begin{equation} \mathcal{L}_{cmd} = \sum_{i=1}^4 \sum_{j=1}^{C_i} \mathcal{S}\left(g_d^{i,j}, g_r^{i,j}\right) \log\left(\frac{\mathcal{S}\left(g_d^{i,j}, g_r^{i,j}\right)}{\mathcal{S}\left(f_d^{i,j}, f_r^{i,j}\right)}\right), \end{equation}
% where $\mathcal{S}(x,y)=\frac{x\cdot y}{||x|| ||y||}$ denotes cosine similarity between feature vectors $x$ and $y$. 
The distillation of cross-modal knowledge between student features \(\{g_r^i, g_d^i\}_{i=1}^4\) and teacher features \(\{f_r^i, f_d^i, f_e^i, f_l^i\}_{i=1}^4\) is achieved through:
\begin{equation}
\mathcal{L}_{cmd} = \sum_{i=1}^4 \sum_{j=1}^{C_i} \tilde{\mathcal{S}}\left(g_d^{i,j}, g_r^{i,j}\right) \log\left(\frac{\tilde{\mathcal{S}}\left(g_d^{i,j}, g_r^{i,j}\right)}{\tilde{\mathcal{S}}\left(f_d^{i,j}, f_r^{i,j}\right)}\right),
\end{equation}
where \(\mathcal{S}(x,y) = \frac{x \cdot y}{\|x\| \|y\|}\) denotes the cosine similarity between feature vectors \(x\) and \(y\), and \(\tilde{\mathcal{S}}(x, y) = \frac{\mathcal{S}(x, y) + 1}{2}\) is its \textbf{\textit{normalized form within \([0, 1]\)}} to ensure numerical stability for the logarithmic operation.
To further mitigate potential instability caused by negative cosine values—especially when the similarities of teacher and student features differ in sign—we \textit{average the similarity scores across batch samples with the same feature shape}. Empirically, this averaging leads to \textit{non-negative values} when modalities are semantically aligned. Nonetheless, we adopt the normalized similarity \(\tilde{\mathcal{S}}\) to guarantee theoretical robustness. 
This formulation aligns cross-modal representations by minimizing the discrepancy between teacher and student similarities across modalities. After applying cross-modal correspondence distillation, the inherent unimodal bias in this task—as well as the bias introduced by unimodal knowledge distillation—are largely mitigated, as shown in Tab.~\ref{tab:ablation_loss_umd}.

% \noindent \textbf{Multimodal Distillation.} Based on $\mathcal{L}_{umd}$ and $\mathcal{L}_{cmd}$, we have distilling the unimodal and cross-modal knowledge from the strong teacher $\mathcal{F}_{ma}$. To further utilize the knowledge in $\mathcal{F}_{ma}$, we also introduce a multimodal distillation which means that fusing all the features from both teacher and student, then impose loss functions between the fused features. Unlike ~\cite{chen2021spatial} which introduces additional layers for balancing the multimodal features, we directly average the features to get the fused ones to avoid adding extra training parameters while maintaining all the feature characteristics from the segmentation backbone. The fused features are obtained as:
% \begin{equation}
% \{f_{f}^i\}_{i=1}^4=\mathcal{A}\{f_r^i, f_d^i, f_e^i, f_l^i\}_{i=1}^4,
% \{g_{f}^i\}_{i=1}^4=\mathcal{A}\{g_r^i, g_d^i,\}_{i=1}^4,
% \end{equation}
% where $A$ denotes the mean operation and $\{f_{f}^i\}_{i=1}^4$ and $\{g_{f}^i\}_{i=1}^4$ stand for the fused features for teacher and student models. The multimodal distillation is performed between the fused features with KL divergence loss:
% \begin{equation}
%     \mathcal{L}_{MMD} = KL(\{g_{f}^i\}_{i=1}^4,\{f_{f}^i\}_{i=1}^4).
% \end{equation}
\begin{figure*}[t!]
    \centering
    \includegraphics[width=\textwidth]{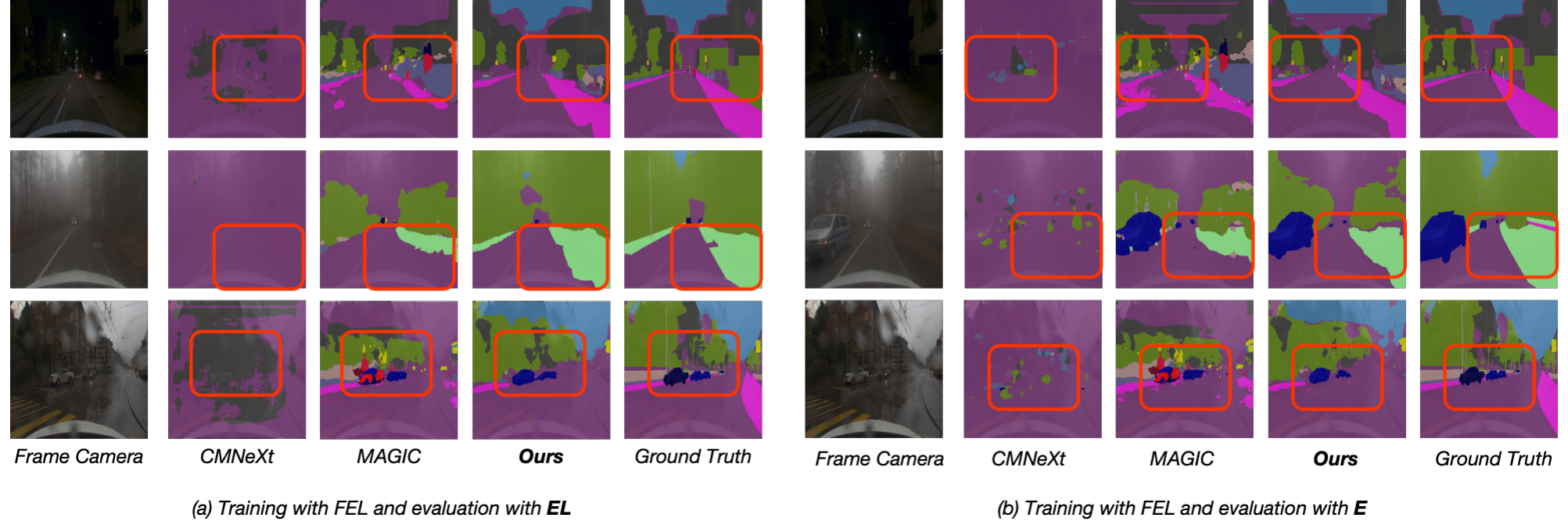}
    % \vspace{-20pt}
    \caption{Qualitative comparison on MUSES.
    }
    \vspace{-8pt}
    \label{fig:visual}
\end{figure*}

\begin{table*}[t!]
\renewcommand{\tabcolsep}{8pt}
\caption{Results of anymodal semantic segmentation validation with three modalities (F: frame camera, E: event cameras, L: LiDAR sensor) on real-world MUSES with SegFormer-B0.}
\resizebox{\linewidth}{!}{
\begin{tabular}{c|c|c|ccccccc|c}
\midrule
\multirow{2}{*}{Method} & \multirow{2}{*}{Pub.} & \multirow{2}{*}{Training} & \multicolumn{7}{c}{Anymodal Evaluation} & \multirow{2}{*}{Mean}  \\ \cmidrule{4-10}
 &  &  & F & E & L & FE & FL & EL & FEL &  \\ \midrule
CMX~\cite{zhang2023cmx} & T-ITS 2023 & \multirow{3}{*}{} & 2.52 & 2.35 & 3.01 & 41.15 & 41.25 & 2.56 & 42.27 & 19.30 \\ \cmidrule{1-2} \cmidrule{4-11} 
CMNeXt~\cite{zhang2023delivering} & CVPR 2023 & \multirow{3}{*}{FEL} & 3.50 & 2.77 & 2.64 & 6.63 & 10.28 & 3.14 & 46.66 & 10.80 \\ \cmidrule{1-2} \cmidrule{4-11} 
MAGIC~\cite{zheng2024centering} & ECCV 2024 &  & 43.22 & 2.68 & \underline{22.95} & 43.51 & 49.05 & \underline{22.98} & 49.02 & 33.34 \\ \cmidrule{1-2} \cmidrule{4-11} 
Any2Seg~\cite{zheng2024learning} & ECCV 2024 &  & \underline{44.40} & \underline{3.17} & 22.33 & 44.51 & \underline{49.96} & 22.63 & \underline{50.00} & \underline{33.86} \\ \cmidrule{1-2} \cmidrule{4-11} 
 Ours & - & & \textbf{46.01} & \textbf{19.57} & \textbf{32.13} & \textbf{46.29} & \textbf{51.25} & \textbf{35.21} & \textbf{51.14} & \textbf{40.23}  \\ \midrule
\textit{w.r.t} SoTA & - & - & \textbf{+1.61} & \textbf{+16.40} & \textbf{+9.80} & \textbf{+1.78} & \textbf{+1.29} & \textbf{+12.58} & \textbf{+1.14} & \textbf{+6.37} \\
\bottomrule
\end{tabular}}
\label{Tab:MUSES}
% \vspace{-12pt}
\end{table*}
\subsection{Modality-agnostic Distillation}
After addressing the unimodal bias problem in the representation spaces, we also focus on transferring task-related semantic information at the prediction level for further utilization of the pre-trained knowledge in the teacher model. Specifically, the segmentation maps predicted by the multimodal teacher \( P_{mm} \) are used as supervision signals for the predictions of the anymodal student segmentor \( P_{am} \).
The modality-agnostic distillation loss is formulated as:
\begin{equation} 
\mathcal{L}_{mad} = \frac{1}{N} \sum_{i=1}^{N} \sum_{k=1}^{K} P_{am}^{i,k} \log\left(\frac{P_{am}^{i,k}}{P_{mm}^{i,k}}\right). 
\end{equation}
Additionally, there is also a supervised loss imposed between the $P_{am}$ and the GT:
\begin{equation} 
\mathcal{L}_{sup} = -\frac{1}{N} \sum_{i=1}^{N} \sum_{k=1}^{K} y^i_k \log\left(P_{am}^{i,k}\right). 
\end{equation}

The total loss for training the anymodal student segmentor combines the supervised objective with multiple distillation terms, formulated as:
\begin{equation}
\mathcal{L}_{\text{total}} = \mathcal{L}_{sup} + \lambda_{\text{mad}} \mathcal{L}_{mad} + \alpha \mathcal{L}_{umd} + \beta \mathcal{L}_{cmd},
\end{equation}
where \( \lambda_{\text{mad}} \), \( \alpha \), and \( \beta \) are weighting coefficients for the modality-adaptive, unimodal, and cross-modal distillation losses, respectively. The supervised loss \( \mathcal{L}_{sup} \) serves as the primary training signal and is left unweighted for clarity, with all distillation losses scaled relative to it.

\begin{table*}[t!]
\renewcommand{\tabcolsep}{1pt}
\caption{Results of anymodal semantic segmentation validation with four modalities (R: RGB, D: Depth, E: Event, L: LiDAR) on DELIVER using SegFormer-B0 as backbone.}
\resizebox{\linewidth}{!}{
\begin{tabular}{c|ccccccccccccccc|c}
\midrule
\multirow{2}{*}{Method} & \multicolumn{15}{c}{Anymodal Evaluation} & \multirow{2}{*}{Mean}\\ \cmidrule{2-16}
 & R & D & E & L & RD & RE & RL & DE & DL & EL & RDE & RDL & REL & DEL & RDEL & \\ \midrule
CMNeXt~\cite{zhang2023delivering} & 0.86 & 0.49 & \underline{0.66} & 0.37 & 47.06 & 9.97 & 13.75 & 2.63 & 1.73 & \underline{2.85} & 59.03 & 59.18 & 14.73 & \textbf{59.18} & 39.07 & 20.77 \\ \midrule
MAGIC~\cite{zheng2024centering} & \underline{32.60} & \textbf{55.06} & 0.52 & \underline{0.39} & \textbf{63.32} & \underline{33.02} & \underline{33.12} & \textbf{55.16} & \textbf{55.17} & 0.26 & \textbf{63.37} & \textbf{63.36} & \underline{33.32} & \underline{55.26} & \textbf{63.40} & 40.49 \\ \midrule
Ours & \textbf{47.11} & \underline{52.17} & \textbf{17.33} & \textbf{19.01} & \underline{60.37} & \textbf{47.49} & \textbf{48.13} & \underline{52.82} & \underline{52.29} & \textbf{21.47} & \underline{60.16} & \underline{60.60} & \textbf{47.98} & 52.44 & \underline{60.26} & \textbf{46.64} \\ \midrule
\textit{w.r.t} SoTA & \textbf{+14.51} & -2.89 & \textbf{+16.81} & \textbf{+18.62} & -2.95 & \textbf{+14.47} & \textbf{+15.01} & -2.34 & -2.88 & \textbf{+21.21} & -3.21 & -2.76 & \textbf{+14.66} & -2.82 & -3.14 & \textbf{+6.15} \\ 
\bottomrule
\end{tabular}}
\label{Tab:DELIVER}
\vspace{-8pt}
\end{table*}

\begin{table*}[t!]
\centering
\renewcommand{\tabcolsep}{1pt}
\caption{Ablation study of different loss combinations on MUSES dataset~\cite{brodermann2024muses}.}
\resizebox{\linewidth}{!}{
\begin{tabular}{l|ccccccccccccccccc}
\toprule
Loss Combination & F & \textbf{$\Delta \uparrow$} & E & \textbf{$\Delta \uparrow$} & L & \textbf{$\Delta \uparrow$} & FE & \textbf{$\Delta \uparrow$} & FL & \textbf{$\Delta \uparrow$} & EL & \textbf{$\Delta \uparrow$} & FEL & \textbf{$\Delta \uparrow$} & Mean & \textbf{$\Delta \uparrow$} \\ \midrule
$\mathcal{L}_{sup}$ & 43.69 & - &  22.35 & - & 32.14 & - & 44.58 & - & 48.53 & - & 35.40 & - & 48.35 & - & 39.29 & -  \\ \midrule
$\mathcal{L}_{sup}$ + $\lambda \mathcal{L}_{mad}$   & 43.71 & +0.02 & 23.00 & +0.65 & 34.70 & +2.56 & 44.18 & -0.40 & 49.13 & +0.60 & 37.23 & +1.83 & 48.79 & +0.44 & 40.11 & +0.82 \\ \midrule
$\mathcal{L}_{sup}$ + $\lambda \mathcal{L}_{mad}$ + $\alpha \mathcal{L}_{umd}$ & 45.82 & +2.13 & 19.26 & -3.09 & 31.79 & -0.35 & 45.88 & +1.30 & 51.11 & +2.58 & 33.56 & -1.84 & 50.60 & +0.43 & 39.72 & +0.43 \\ \midrule
$\mathcal{L}_{sup}$ + $\lambda \mathcal{L}_{mad}$ + $\alpha \mathcal{L}_{umd}$ + $\beta \mathcal{L}_{cmd}$ & 46.01 & +2.32 & 19.57 & -2.78 & 32.13 & -0.01 & 46.29 & +1.71 & 51.25 & +2.72 & 35.21 & -0.21 & 51.14 & +2.79 & 40.23 & +0.94 \\ 
\bottomrule
\end{tabular}
}
\label{tab:loss_combine}
\vspace{-8pt}
\end{table*}

\begin{table*}[t!]
\centering
\renewcommand{\tabcolsep}{4pt}
\caption{Ablation study on the effect of different parameters for $L_{mad}$ in our framework on MUSES dataset~\cite{brodermann2024muses}. More results can be found in Table~\ref{tab:ablation_loss_mad_suppl}.}
\resizebox{\linewidth}{!}{
\begin{tabular}{cccccccccccccccccc}
\toprule
$\lambda$ & F & \textbf{$\Delta \uparrow$} & E & \textbf{$\Delta \uparrow$} & L & \textbf{$\Delta \uparrow$} & FE & \textbf{$\Delta \uparrow$} & FL & \textbf{$\Delta \uparrow$} & EL & \textbf{$\Delta \uparrow$} & FEL & \textbf{$\Delta \uparrow$} & Mean & \textbf{$\Delta \uparrow$} \\ \midrule
1   & 43.97 & - & 22.33 & - & 31.90 & - & 44.82 & - & 48.61 & - & 35.14 & - & 48.33 & - & 39.30 & - \\ \midrule
% 10   & 43.84 & -0.13 & 23.21 & +0.88 & 32.71 & +0.81 & 44.08 & -0.74 & 49.16 & 0.55 & 34.97 & -0.17 & 48.08 & -0.25 & 39.44 & +0.14 \\ \midrule
20   & 44.08 & +0.11 & 22.76 & +0.43 & 32.35 & +0.45 & 44.37 & -0.45 & 49.33 & +0.72 & 34.73 & -0.41 & 48.79 & +0.46 & 39.49 & +0.19 \\ \midrule
 \textbf{50}   & 43.71 & -0.26 & 23.00 & +0.67 & \textbf{34.70} & +2.80 & 44.18 & -0.64 & 49.13 & +0.52 & \textbf{37.23} & +2.09 & \textbf{48.79} & +0.46 & \textbf{40.11} & \textbf{+0.81} \\ \midrule
% 60   & 44.02 & +0.05 & 22.74 & +0.41 & 33.82 & +1.92 & 44.29 & -0.53 & 49.36 & +0.75 & 36.69 & +1.55 & 48.54 & +0.21 & 39.92 & +0.62 \\ \midrule
80   & 43.84 & -0.13 & 22.86 & +0.53 & 33.78 & +1.88 & 44.25 & -0.57 & 49.43 & +0.82 & 36.57 & +1.43 & 48.72 & +0.39 & 39.92 & +0.62 \\ 
% \midrule
% 100   & 43.75 & -0.22 & 22.87 & +0.54 & 34.00 & \cellcolor{orange!30}+2.10 & 44.17 & -0.65 & 49.36 & +0.75 & 36.60 & +1.46 & 48.64 & +0.31 & 39.91 & +0.61 \\ 
\bottomrule
\end{tabular}
}
\label{tab:ablation_loss_mad}
\vspace{-8pt}
\end{table*}

\section{Experiments}
\noindent \textbf{Experimental Setup.} We evaluate our method on synthetic and real-world multi-sensor datasets. The MUSES dataset~\cite{brodermann2024muses}, recorded in Switzerland, includes driving sequences designed to address challenges from adverse visual conditions. It features multi-sensor data from a high-resolution frame camera, an event camera, and MEMS LiDAR, enhancing annotation quality and supporting robust multimodal semantic segmentation. Each sequence is annotated with 2D panoptic labels for accurate ground truth. The DELIVER dataset~\cite{zhang2023delivering} includes RGB, depth, LiDAR, and event data across 25 semantic categories, covering various environmental conditions and sensor failures for thorough evaluations. We follow the official data processing and split protocols. More implementation details can be found in Sec.~\ref{appendix:implementation_detail}

\subsection{Results}
As shown in Tab.~\ref{Tab:MUSES}, our method achieves the highest mIoU of \textbf{40.23}, surpassing all state-of-the-art (SoTA) baselines. CMX~\cite{zhang2023cmx} and CMNeXt~\cite{zhang2023delivering} perform poorly (mIoU: 19.30\% and 10.80\%), due to over-reliance on RGB. Similarly, MAGIC~\cite{zheng2024centering} and Any2Seg~\cite{zheng2024learning} depend heavily on depth, leading to large performance drops when it's absent—exposing their unimodal bias.
In contrast, our method shows strong, balanced performance across all modalities: RGB (46.01\%), Event (19.57\%), and LiDAR (32.13\%), as well as paired combinations: FE (46.29\%), FL (51.25\%), EL (35.21\%), and FEL (51.14\%), demonstrating robust cross-modal learning. Significant gains in Event (+16.40\%) and FL (+12.58\%) further highlight the ability to handle challenging or sparse modalities. Fig.~\ref{fig:visual} provides qualitative comparisons.
Tab.~\ref{Tab:DELIVER} reports results on the synthetic DELIVER benchmark using SegFormer-B0. Our method achieves a top mIoU of \textbf{46.64\%}, outperforming MAGIC~\cite{zheng2024centering} by \textbf{+6.15\%}. On individual modalities, it achieves 47.11\% (R), 52.17\% (D), and 19.01\% (L), with +14.51\% (R) and +18.62\% (L) gains over MAGIC. Event and LiDAR improvements further confirm robustness to unimodal bias.
For paired modalities, RD, RE, and RL achieve mIoUs of 60.37\%, 47.49\%, and 48.13\%, with notable gains over MAGIC (+14.47\% RE, +15.01\% RL). These results demonstrate our model's strength in capturing cross-modal dependencies while remaining resilient to missing inputs.
Overall, the results confirm the effectiveness of our method in both unimodal and multimodal settings, offering strong generalization across diverse sensor combinations.

\begin{table*}[t!]
\centering
\renewcommand{\tabcolsep}{5pt}
\caption{Ablation on different parameters for $L_{umd}$ in our framework on MUSES dataset~\cite{brodermann2024muses}.}
\resizebox{\textwidth}{!}{
\begin{tabular}{cccccccccccccccccc}
\toprule
$\alpha$ & F & \textbf{$\Delta \uparrow$} & E & \textbf{$\Delta \uparrow$} & L & \textbf{$\Delta \uparrow$} & FE & \textbf{$\Delta \uparrow$} & FL & \textbf{$\Delta \uparrow$} & EL & \textbf{$\Delta \uparrow$} & FEL & \textbf{$\Delta \uparrow$} & Mean & \textbf{$\Delta \uparrow$} \\ \midrule
w/o & 43.71 & - & 23.00 & - & 34.70 & - & 44.18 & - & 49.13 & - & 37.23 & - & 48.79 & - & 40.11 & - \\ \midrule
3 & 45.38 & +1.67 & 20.64 & -2.36 & 31.37 & -3.33 & 45.43 & +1.25 & 50.53 & +1.40 & 33.65 & -3.58 & 49.93 & +1.14 & 39.56 & -0.55 \\ \midrule
\textbf{5} & 45.82 & +2.11 & 19.26 & -3.74 & 31.79 & -2.91 & 45.88 & +1.70 & 51.11 & +1.98 & 33.56 & -3.67 & 50.60 & +1.81 & 39.72 & -0.39\\ \midrule
7 & 46.09 & +2.38 & 17.84 & -5.16 & 31.81 & -2.89 & 46.18 & +2.00 & 51.36 & +2.23 & 33.43 & -3.80 & 51.01 & +2.22 & 39.67 & -0.44 \\ \midrule
10 & 46.17 & +2.46 & 15.74 & -7.26 & 31.95 & -2.75 & 46.37 & +2.19 & 51.17 & +2.04 & 33.26 & -3.97 & 51.08 & +2.29 & 39.39 & -0.72 \\
\bottomrule
\end{tabular}
}
\label{tab:ablation_loss_umd}
\vspace{-8pt}
\end{table*}

\begin{table*}[t!]
\centering
\caption{Ablation on different parameters for add $L_{cmd}$ with $L_{umd}$ on MUSES dataset~\cite{brodermann2024muses}. w/o means the framework is only trained with $L_{mad}$ + $L_{umd}$. More results can be found in Table.~\ref{tab:ablation_loss_cmd_baseumd_suppl}.}
\setlength{\tabcolsep}{4pt}
\resizebox{\textwidth}{!}{
\begin{tabular}{cccccccccccccccccc}
\toprule
$\beta$ & F & \textbf{$\Delta \uparrow$} & E & \textbf{$\Delta \uparrow$} & L & \textbf{$\Delta \uparrow$} & FE & \textbf{$\Delta \uparrow$} & FL & \textbf{$\Delta \uparrow$} & EL & \textbf{$\Delta \uparrow$} & FEL & \textbf{$\Delta \uparrow$} & Mean & \textbf{$\Delta \uparrow$} \\ \midrule
w/o & 45.82 & - & 19.26 & - & 31.79 & - & 45.88 & - & 51.11 & - & 33.56 & - & 50.60 & - & 39.72 & - \\ \midrule
1 & 45.93 & +0.11 & 18.76 & -0.50 & 31.84 & +0.05 & 45.96 & +0.08 & 51.22 & +0.11 & 33.49 & -0.07 & 50.82 & +0.22 & 39.72 & 0.00 \\ \midrule
3 & 46.04 & +0.22 & 17.74 & -1.52 & 31.42 & -0.37 & 46.08 & +0.20 & 51.27 & +0.16 & 33.46 & -0.10 & 50.99 & +0.39 & 39.57 & -0.15 \\ \midrule
5 & 46.19 & +0.37 & 17.27 & -1.99 & 31.03 & -0.76 & 46.27 & +0.39 & 51.34 & +0.33 & 33.40 & -0.16 & 51.05 & +0.45 & 39.51 & -0.21 \\ \midrule
7 & 46.21 & +0.39 & 17.40 & -1.86 & 31.06 & -0.73 & 46.37 & +0.49 & 51.29 & +0.18 & 33.79 & +0.23 & 51.12 & -0.12 & 39.60 & -0.12 \\ \midrule
\textbf{10} & 46.01 & +0.19 & 19.57 & +0.31 & 32.13 & +0.34 & 46.29 & +0.41 & 51.25 & +0.14 & 35.21 & +1.65 & 51.14 & +0.54 & 40.23 & +0.51 \\ \midrule
13 & 46.57 & +0.75 & 18.24 & -1.02 & 30.88 & -0.91 & 46.74 & +0.86 & 51.09 & -0.02 & 33.88 & +0.32 & 50.76 & +0.16 & 39.74 & +0.02 \\
\midrule
15 & 46.03 & +0.21 & 14.10 & -8.90 & 31.12 & -3.58 & 45.99 & +1.81 & 50.97 & +1.84 & 31.42 & -5.81 & 50.49 & -0.11 & 38.59 & -1.13 \\
\midrule
20 & 45.95 & +0.13 & 15.19 & -7.81 & 30.61 & -4.09 & 45.80 & +1.62 & 51.19 & +2.06 & 30.55 & -6.68 & 50.41 & +1.62 & 38.53 & -1.58 \\
\bottomrule
\end{tabular}
}
\label{tab:ablation_loss_cmd_baseumd}
\vspace{-12pt}
\end{table*}

% \begin{figure*}[ht!]
%     \centering
%     \includegraphics[width=\textwidth]{images/UMD_CMD.pdf}
%     % \vspace{-16pt}
%     \caption{Hyper-parameter selection for $L_{umd}$ and $L_{cmd}$.
%     }
%     % \vspace{-12pt}
%     \label{fig:umd_cmd}
% \end{figure*}

\section{Ablation Study}
% \noindent \textbf{Effectiveness of Loss Functions}
% Tab.~\ref{tab:loss_combine} shows results of different loss function combinations on the MUSES dataset~\cite{brodermann2024muses}. The baseline using only the supervised loss (\( \mathcal{L}_{sup} \)) achieves a mean mIoU of 39.29\%. Adding the modality-adaptive loss (\( \lambda \mathcal{L}_{mad} \)) improves performance, with gains in paired and combined modalities.  
% Adding the unimodal-distillation loss (\( \alpha \mathcal{L}_{umd} \)) to \( \mathcal{L}_{sup} \) + \(\lambda \mathcal{L}_{mad} \) results in more significant gains. Notable improvements include RGB (F: +2.13\%), FE (+1.30\%), and FL (+2.58\%), highlighting the value of leveraging unimodal features.  
% The full loss combination, incorporating the cross-modal distillation loss (\( \beta \mathcal{L}_{cmd} \)), achieves the highest mean mIoU of 40.23\%, with consistent improvements across all modalities. Paired modalities FL (+2.72\%) and FEL (+2.79\%) show the largest gains, while Event (E) improves slightly (+0.43\%). These results demonstrate the method's robustness in handling complex multimodal interactions. Overall, each loss component plays a critical role in improving anymodal segmentation, with the full combination proving most effective in leveraging multimodal data for real-world scenarios.
\noindent \textbf{Effectiveness of Loss Functions} 
Tab.~\ref{tab:loss_combine} reports results of different loss combinations on the MUSES dataset~\cite{brodermann2024muses}. Using only supervised loss (\( \mathcal{L}_{sup} \)) yields a mean mIoU of 39.29\%. Adding \( \lambda \mathcal{L}_{mad} \) improves performance across paired and combined modalities.
Further including \( \alpha \mathcal{L}_{umd} \) brings notable gains, especially in F (+2.13\%), FE (+1.30\%), and FL (+2.58\%), confirming the benefit of unimodal knowledge distillation.
The full combination with \( \beta \mathcal{L}_{cmd} \) achieves the best performance (mean mIoU \textbf{40.23\%}), with broad improvements, particularly in FL (+2.72\%) and FEL (+2.79\%). Even E improves slightly (+0.43\%), showing robustness across modalities.
In sum, each loss contributes to better segmentation, and the full loss design is most effective for capturing complex multimodal interactions.

\noindent \textbf{Ablation on Teacher Model and Fusion Strategy} 
We compare our PML with fusion-based teacher models MAGIC and CMNeXt to assess their impact on student performance. Unlike CMNeXt's fixed fusion architecture, PML enables both unimodal and cross-modal distillation via parallel modality learning, providing more diverse and semantically aligned supervision. 
\begin{wraptable}{r}{8cm}
\centering
\caption{Ablation on teacher model selection. Results reported as mIoU (\%).}
\label{ab:teachermodel}
\setlength{\tabcolsep}{2pt}
\resizebox{\linewidth}{!}{
\begin{tabular}{lcccccccc}
\toprule
\textbf{Teacher} & \textbf{F} & \textbf{E} & \textbf{L} & \textbf{FE} & \textbf{FL} & \textbf{EL} & \textbf{FEL} & \textbf{Mean} \\
\midrule
MAGIC      & 43.87 & 13.22 & 30.90 & 43.91 & 48.41 & 33.68 & 47.75 & 37.39 (-2.84) \\
CMNeXt     & 43.80 & 10.79 & 23.15 & 48.34 & 43.97 & 33.52 & 47.64 & 35.89 (-4.34) \\ \midrule
PML (Ours) & 46.01 & 19.57 & 32.13 & 46.29 & 51.25 & 35.21 & 51.14 & \textbf{40.23} \\
\bottomrule
\end{tabular}}
\vspace{-8pt}
\end{wraptable}
Results in Tab.~\ref{ab:teachermodel} show that distilling from fused features, as in MAGIC and CMNeXt, degrades performance. CMNeXt yields the largest drop (–4.34\% mIoU), indicating that rigid fusion limits generalization. In contrast, PML achieves the best performance by preserving both modality-specific and shared representations.
% While alternative fusion strategies are worth exploring, t
These results confirm PML’s flexibility and effectiveness as a general-purpose teacher for anymodal segmentation.

\noindent \textbf{Ablation on Hyper-Parameter Selection} 
We study the effect of hyper-parameters \( \lambda \), \( \alpha \), and \( \beta \) for \( \mathcal{L}_{mad} \), \( \mathcal{L}_{umd} \), and \( \mathcal{L}_{cmd} \), respectively (Tab.~\ref{tab:ablation_loss_mad}–\ref{tab:ablation_loss_cmd_baseumd}).
Increasing \( \lambda \) improves performance up to a point, particularly benefiting underrepresented modalities, before plateauing due to diminishing returns. Varying \( \alpha \) reveals that moderate values enhance unimodal contributions, while overly large values lead to degradation. Adding \( \mathcal{L}_{cmd} \) with an appropriate \( \beta \) further improves results, but excessive weighting can suppress unimodal learning.

\begin{table*}[t!]
\centering
\caption{Discussion study on the effect of performing KD with fused features on MUSES dataset~\cite{brodermann2024muses}. w/o means the framework is only trained with $L_{mad}$ + $L_{umd}$ + $L_{cmd}$. 
% More results can be found in Table.~\ref{tab:ablation_loss_cmd_umd_mmd_suppl} in the appendix.
}
\setlength{\tabcolsep}{2.5pt}
\resizebox{\textwidth}{!}{
\begin{tabular}{cccccccccccccccccc}
\toprule
$\lambda$ & F & \textbf{$\Delta \uparrow$} & E & \textbf{$\Delta \uparrow$} & L & \textbf{$\Delta \uparrow$} & FE & \textbf{$\Delta \uparrow$} & FL & \textbf{$\Delta \uparrow$} & EL & \textbf{$\Delta \uparrow$} & FEL & \textbf{$\Delta \uparrow$} & Mean & \textbf{$\Delta \uparrow$} \\ \midrule
w/o & 46.01 & - & 19.57 & - & 32.13 & - & 46.29 & - & 51.25 & - & 35.21 & - & 51.14 & - & 40.23 & - \\ \midrule
1 & 46.32 & +0.31 & 17.99 & -1.58 & 31.38 & -0.75 & 46.80 & +0.51 & 51.01 & -0.24 & 33.92 & -1.29 & 50.95 & -0.19 & 39.77 & -0.46 \\ \midrule
3 & 46.34 & +0.33 & 17.27 & -2.30 & 31.36 & -0.77 & 46.81 & +0.52 & 51.20 & -0.05 & 33.87 & -1.34 & 51.13 & -0.01 & 39.71 & -0.52 \\ 
\bottomrule
\end{tabular}
}
\label{tab:ablation_loss_cmd_umd_mmd}
\vspace{-12pt}
\end{table*}

\noindent \textbf{Rationality of Unimodal Distillation}
The effectiveness of \( \mathcal{L}_{umd} \) is demonstrated in Tab.~\ref{tab:ablation_loss_umd}. Adding \( \mathcal{L}_{umd} \) with \( \alpha = 1 \) improves single-modality performance, particularly for RGB (\( F: +0.83\% \)) and paired modalities such as FL (\(+0.42\%\)). The best performance for single-modality tasks is observed at \( \alpha = 10 \), where the mean mIoU for RGB increases to \textbf{46.17} (\(+2.46\% \)), and FL achieves \textbf{51.17\%} (\(+2.19\%\)). Obviously, \( \mathcal{L}_{umd} \) effectively improves single-modality segmentation performance by facilitating knowledge transfer within each modality. However, the results also highlight a trade-off: while unimodal distillation enhances performance for individual modalities, it may compromise the model's ability to handle complex multimodal combinations. Careful tuning of \( \alpha \) is therefore critical to achieving a balance between single-modality and multimodal segmentation performance.

\noindent \textbf{Rationality of Cross-modal Distillation.}  
Unimodal knowledge distillation improves segmentation performance on RGB images but introduces unimodal bias, reducing performance on other modalities. This bias arises as the model over-relies on RGB, which is easier to learn, limiting its generalization across diverse inputs. As shown in Tab.~\ref{tab:ablation_loss_umd}, performance on Event (-3.74\%) and LiDAR (-2.91\%) modalities declines significantly. While unimodal knowledge transfer enhances single-modality performance, it negatively impacts multimodal performance in mixed-modality scenarios.
Cross-modal correspondence distillation mitigates this bias by aligning representations across modalities, improving segmentation performance on diverse inputs. It balances the trade-off between single-modality and multimodal performance, emphasizing the need to carefully tune \( \beta \) for optimal results.
However, larger \( \beta \) values (\( \beta > 13 \)) degrade performance, particularly on Event (-8.90\%) and LiDAR (-3.58\%), suggesting that overemphasizing cross-modal distillation can overshadow individual modality learning, leading to performance trade-offs.

\noindent \textbf{Why not doing KD between Fused Features?}
To explore the possibility of transferring knowledge between fused features from teacher to student models, we conduct experiments on the MUSES dataset, leveraging the fusion method described in~\cite{zheng2024centering}. The results are summarized in Tab.~\ref{tab:ablation_loss_cmd_umd_mmd}. It is evident that applying knowledge distillation (KD) directly on fused features causes a notable performance drop across most metrics.
For instance, compared to the baseline without KD on fused features (w/o), which achieves a mean mIoU of 40.23\%, all settings with KD exhibit reduced performance. At \( \lambda = 1 \), the mean mIoU drops to 39.77\% (-0.46\%), while higher \( \lambda \) values exacerbate the decline, with \( \lambda = 10 \) resulting in a mean mIoU of 39.37\% (-0.86\%). These results highlight the inefficacy of such distillation methods.
Examining individual modalities further reveals this trend. For the Event (E) and LiDAR (L) modalities, performance consistently degrades as \( \lambda \) increases, with Event mIoU decreasing by up to -3.96\% at \( \lambda = 10 \). Paired and fused modalities also show minimal improvement or slight degradation, such as FEL achieving a marginal gain of +0.03\% at \( \lambda = 5 \) but regressing at \( \lambda = 10 \).
The results suggest that directly applying knowledge distillation between fused features fails to effectively guide the student model. Instead of transferring meaningful knowledge, the fused features introduce noise and misalignment, limiting their utility in this context. This reinforces the importance of designing specialized mechanisms for knowledge distillation, targeting individual or structured features rather than indiscriminately fused representations.

\begin{figure}[t!]
    \centering
    \includegraphics[width=0.95\linewidth]{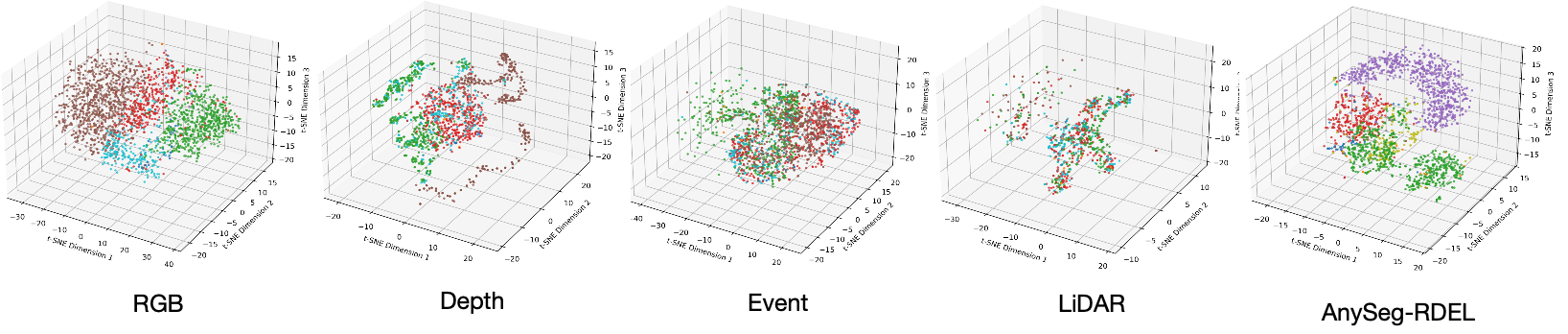}
    \vspace{-8pt}
    \caption{TSNE visualization of multi-modal features (RGB-R, Depth-D, Event-E, and LiDAR-L) and the learned features of our AnySeg framework.
    }
    \vspace{-16pt}
    \label{fig:TSNE}
\end{figure}

\noindent \textbf{t-SNE Visualization}
Fig.~\ref{fig:TSNE} presents t-SNE visualizations of multimodal features, including RGB, Depth, Event, LiDAR, and the features under our AnySeg framework. Individual modality plots reveal distinct clusters, reflecting semantic separability. RGB and Depth exhibit relatively compact clusters, indicating strong discriminative power, whereas Event and LiDAR show more dispersed and overlapping clusters, highlighting weaker performance when used independently.
The learned feature spaces, particularly AnySeg-RDL and AnySeg-RDEL, show notable improvements in cluster compactness and separability. AnySeg-RDEL, integrating RGB, Depth, Event, and LiDAR, achieves the most coherent and well-separated clusters, demonstrating the framework's robustness in leveraging complementary information across modalities. These results underscore AnySeg's effectiveness in addressing the limitations of individual modalities, achieving robust multi-modal feature representation, and enhancing segmentation performance through cross-modal fusion.
\begin{table}[t!]
\centering
\caption{Performance under different conditions in the DELIVER with RGB-D modalities.}
\label{tab:deliver_conditions_transposed}
\setlength{\tabcolsep}{12pt}
\resizebox{\linewidth}{!}{
\begin{tabular}{lcccccccccccc}
\toprule
\textbf{Metric} & \textbf{All} & \textbf{Cloud} & \textbf{Fog} & \textbf{Night} & \textbf{Rain} & \textbf{Sun} & \textbf{M.B.} & \textbf{O.E.} & \textbf{U.E.} & \textbf{L.J.} & \textbf{E.L.} \\
\midrule
mIoU (\%) & 55.99 & 57.94 & 55.02 & 54.64 & 56.24 & 56.82 & 54.73 & 53.71 & 53.15 & 54.66 & 54.18 \\
mF1  (\%) & 66.97 & 68.80 & 65.33 & 66.16 & 67.43 & 67.46 & 64.88 & 64.59 & 64.10 & 64.51 & 65.04 \\
mAcc (\%) & 64.58 & 65.83 & 64.51 & 63.62 & 64.91 & 64.41 & 63.82 & 61.88 & 62.06 & 62.46 & 62.65 \\
\bottomrule
\end{tabular}}
\vspace{-8pt}
\end{table}

\noindent \textbf{Cross-Sensor Generalization under Scene Degradations.}  
We test AnySeg's performance on the DELIVER using RGB-D under various challenging conditions. As shown in Tab.~\ref{tab:deliver_conditions_transposed}, AnySeg maintains consistent performance across diverse sensor and environmental degradations. The model handles cloud, rain, and sun conditions well, with only minor drops under more severe challenges like motion blur, extreme exposure, and sensor noise (e.g., lidar jitter, event resolution). Despite these disruptions, the model achieves over 53\% mIoU in all cases, demonstrating strong generalization. 
% These results confirm the adaptability of our framework to real-world conditions, where sensor noise and environmental variability are common.

\section{Conclusion}
In this paper, we addressed the challenge of unimodal bias in multimodal semantic segmentation, where reliance on specific modalities leads to performance drops when modalities are missing. We proposed the first framework for anymodal segmentation using unimodal and cross-modal distillation. A PML strategy ensures a strong teacher model, while multiscale distillation transfers feature-level knowledge. By distilling unimodal distributions with cross-modal correspondences, we reduce modality dependency. Additionally, modality-agnostic semantic distillation enables robust prediction-level knowledge transfer. Experiments on synthetic and real multi-sensor benchmarks validate the superior performance of our framework. Furthermore, our discussion on fused feature distillation highlights the need for specialized mechanisms targeting individual or structured features rather than indiscriminate fusion.

\noindent \textbf{Limitations \& Broader Impact}
A key limitation of AnySeg is the additional computational cost introduced by the teacher-student framework for improving missing modality robustness and performance during training.
This work advances multi-modal machine learning, particularly in visual pattern recognition, by addressing unimodal bias in multimodal segmentors. These models often struggle when certain modalities are missing, a common challenge in real-world applications. While our approach has potential societal impacts, we do not identify any specific concerns at this time.

\label{limit}

\bibliographystyle{ieeetr} % 或者 unsrtnat, plainnat
\bibliography{neurips_2025}

%%%%%%%%%%%%%%%%%%%%%%%%%%%%%%%%%%%%%%%%%%%%%%%%%%%%%%%%%%%%

\clearpage
\appendix
\section{Appendix}

\subsection{Implementation Details.} \label{appendix:implementation_detail}
All experiments on MUSES were conducted on 8 NVIDIA 3090 GPUs, while experiments on DELIVER utilized 4 NVIDIA A100 GPUs. The initial learning rate was set to \(6 \times 10^{-5}\) and adjusted using a polynomial decay strategy with a power of 0.9 over 200 epochs. Additionally, a 10-epoch warm-up phase was applied at 10\% of the initial learning rate to stabilize training. The AdamW optimizer was employed, and the batch size was set to 16. Input modality data was cropped to \(1024 \times 1024\) resolution for consistency across benchmarks.

\begin{figure*}[ht!]
    \centering
    \includegraphics[width=0.85\textwidth]{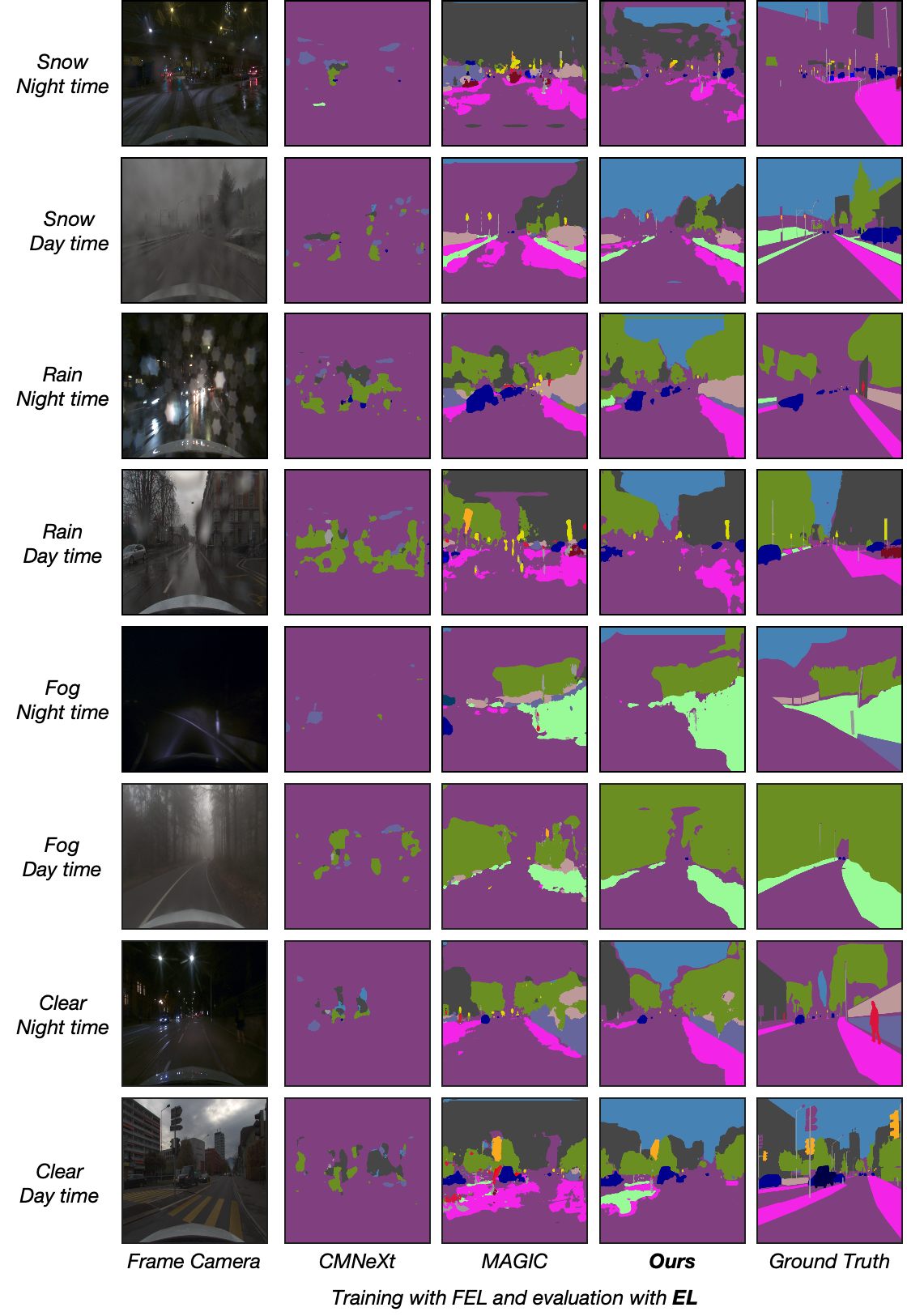}
    \caption{Additional qualitative comparisons that highlight our method’s robustness under diverse dropout conditions.}
    \label{fig:dropout conditions}
\end{figure*}
\begin{figure*}[ht!]
    \centering
    \includegraphics[width=\textwidth]{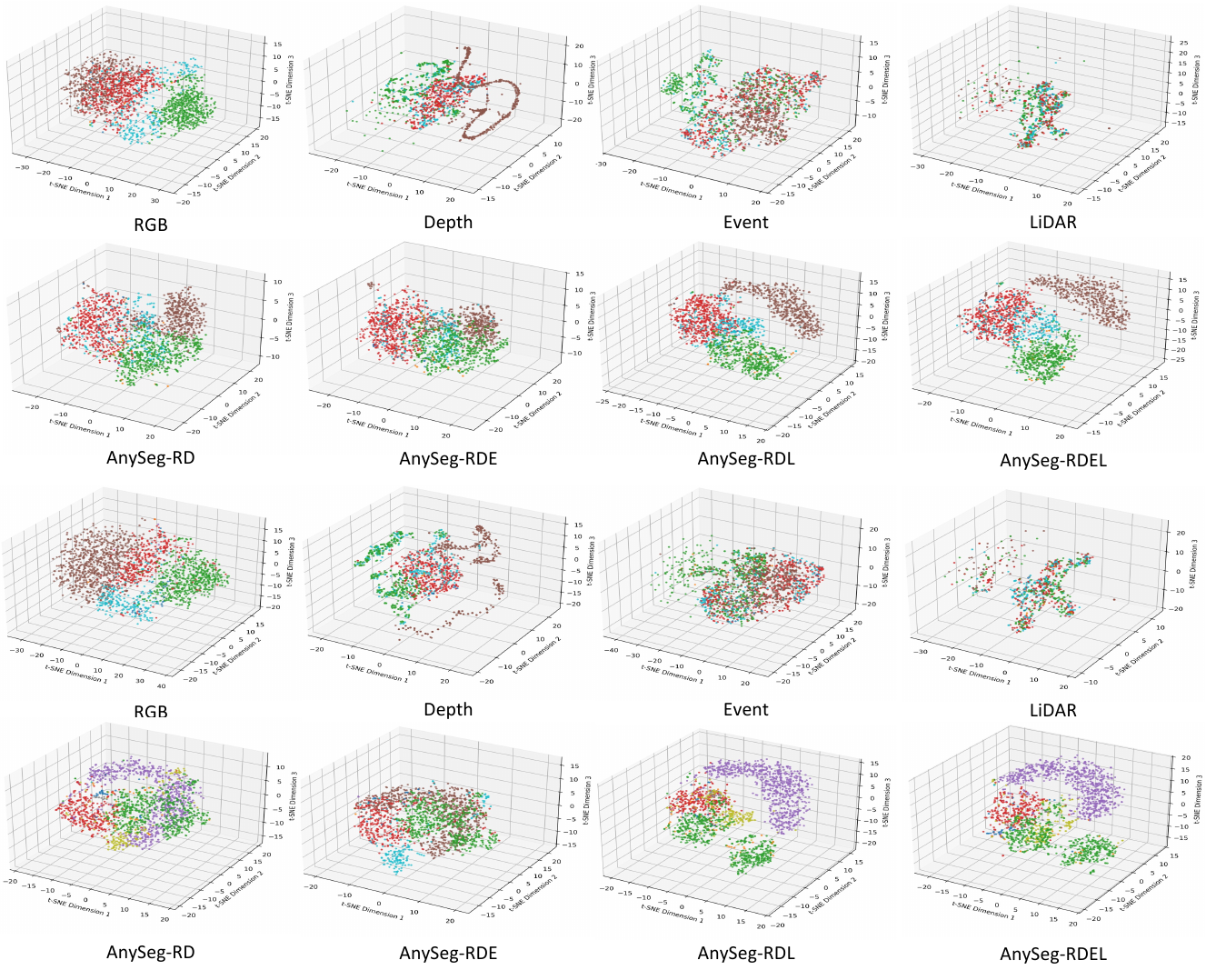}
    \caption{t-SNE visualization of multi-modal features (RGB, Depth, Event, and LiDAR) extracted by the SegFormer backbone and the features of our AnySeg framework.}
    \label{fig:TSNE_suppl}
\end{figure*}
\subsection{Qualitative Feature Visualization}
We provide additional qualitative comparisons that highlight our method’s robustness under diverse dropout conditions in Figure~\ref{fig:dropout conditions}.

Qualitative results are illustrated in Fig.~\ref{fig:TSNE}, which presents the 3D t-SNE visualizations of multimodal feature spaces. The figure includes the individual feature spaces of RGB, Depth, Event, and LiDAR, as well as the fused features obtained through our AnySeg framework. These visualizations provide several key insights:

\noindent \textbf{(I) Distinct Clusters in Individual Modalities:} The 3D t-SNE plots of individual modalities reveal distinct clusters corresponding to semantic classes. However, modalities like Event and LiDAR exhibit greater feature dispersion, reflecting their limited discriminative power when used independently. While RGB and Depth features display relatively better cluster separability, overlaps between certain semantic classes persist, underscoring the challenges of relying solely on individual modalities.

\noindent \textbf{(II) Improved Separability in Paired Modalities:} Feature spaces derived from modality combinations, such as RGB+Depth or RGB+Depth+Event, show marked improvements in cluster separability and compactness when processed through the AnySeg framework. This highlights the benefit of leveraging complementary information across modalities. For instance, integrating RGB and Depth reduces the ambiguities present in individual modalities, yielding more cohesive and distinct clustering.

\noindent \textbf{(III) Robust Fused Feature Space:} The fused feature space of RGB+Depth+Event+LiDAR, as modeled by the AnySeg framework, demonstrates the most compact and well-separated clusters among all configurations. This indicates the framework's effectiveness in integrating multimodal information and enhancing the semantic distinctions between classes. Compared to individual or paired modalities, the fused features provide a more robust representation of the underlying data.

\noindent \textbf{(IV) Mitigation of Ambiguities in Challenging Modalities:} The AnySeg framework effectively addresses the limitations of challenging modalities like Event and LiDAR. Through robust cross-modal fusion mechanisms, it compensates for the weaknesses of these modalities, resulting in consistent and accurate segmentation across diverse scenarios. The framework’s ability to integrate complementary strengths across modalities ensures superior performance in handling complex data distributions.

These qualitative results corroborate the quantitative findings reported in Tab.~\ref{Tab:MUSES} and Tab.~\ref{Tab:DELIVER}. The clear clustering of semantic classes and improved separability in the fused feature space highlight the superiority of the proposed framework in learning robust, multimodal feature representations. Overall, Fig.~\ref{fig:TSNE} visually demonstrates the capability of the AnySeg framework to effectively harness multimodal information for achieving robust semantic segmentation in both individual and multimodal settings.

\subsection{More Experimental Results}
The experimental results in Table~\ref{Tab:DELVIER3M} validate the superiority of our proposed method in modality-agnostic validation on the DELIVER dataset, utilizing three modalities. Across individual modalities, our method achieves the highest performance in RGB (47.44\%) and Event (17.33\%), significantly outperforming MAGIC by margins of +14.48\% and +15.18\%, respectively, while also maintaining competitive results in Depth (52.48\%). For paired modalities, our approach shows strong cross-modal fusion capabilities, achieving substantial improvements in RGB+Event (RE, 47.65\%, +14.40\%) and Depth+Event (DE, 52.61\%, +3.39\%) over MAGIC. Although MAGIC achieves the highest performance in RGB+Depth (RD, 62.52\%) and combined modalities (RDE, 62.49\%), our method performs competitively in these settings with RDE achieving 60.62. Importantly, our method achieves the highest mean mIoU of 48.45\%, surpassing MAGIC by +4.84\% and CMNeXt by +31.21\%, demonstrating superior generalization across diverse modality configurations. These results underscore the robustness of our framework in addressing modality-agnostic challenges, effectively leveraging complementary information across modalities and maintaining balanced performance even in the presence of weaker modalities like Event.

\begin{table*}[]
\renewcommand{\tabcolsep}{12pt}
\caption{Results of modality-agnostic validation with three modalities (R: RGB, D: Depth, E: Event) on DELIVER.}
\resizebox{\linewidth}{!}{
\begin{tabular}{c|c|cccccccc}
\midrule
\multirow{2}{*}{Method} & \multirow{2}{*}{Training} & \multicolumn{7}{c}{Anymodal Evaluation} & \multirow{2}{*}{Mean} \\ \cmidrule{3-9}
 & & R & D & E & RD & RE & DE & RDE &  \\ \midrule
CMNeXt~\cite{zhang2023delivering} & \multirow{3}{*}{RDE} & 2.69 & 0.21 & 0.78 & 48.04 & 6.92 & 2.19 & 59.84 & 17.24 \\ \cmidrule{1-1} \cmidrule{3-10} 
MAGIC~\cite{zheng2024centering} & & \underline{32.96} & \textbf{55.90} & \underline{2.15} & \textbf{62.52} & \underline{33.25} & \textbf{56.00} & \textbf{62.49} & \underline{43.61} \\ \cmidrule{1-1} \cmidrule{3-10} 
Ours & & \textbf{47.44} & \underline{52.48} & \textbf{17.33} & \underline{61.04} & \textbf{47.65} & \underline{52.61} & \underline{60.62} & \textbf{48.45} \\
\midrule
\textit{w.r.t} SoTA & & \textbf{+14.48} & -3.42 & \textbf{+15.18} & -1.48 & \textbf{+14.40} & -3.39 & -1.87 & \textbf{+4.84}\\
\bottomrule
\end{tabular}}
\label{Tab:DELVIER3M}
\end{table*}

\begin{table*}[t!]
\centering
\renewcommand{\tabcolsep}{8pt}
\caption{Ablation study on the effect of different parameters for $L_{mad}$ in our framework on MUSES dataset~\cite{brodermann2024muses}.}
\resizebox{\linewidth}{!}{
\begin{tabular}{cccccccccccccccccc}
\toprule
$\lambda$ & F & \textbf{$\Delta \uparrow$} & E & \textbf{$\Delta \uparrow$} & L & \textbf{$\Delta \uparrow$} & FE & \textbf{$\Delta \uparrow$} & FL & \textbf{$\Delta \uparrow$} & EL & \textbf{$\Delta \uparrow$} & FEL & \textbf{$\Delta \uparrow$} & Mean & \textbf{$\Delta \uparrow$} \\ \midrule
1   & 43.97 & - & 22.33 & - & 31.90 & - & 44.82 & - & 48.61 & - & 35.14 & - & 48.33 & - & 39.30 & - \\ \midrule
10   & 43.84 & -0.13 & 23.21 & +0.88 & 32.71 & +0.81 & 44.08 & -0.74 & 49.16 & +0.55 & 34.97 & -0.17 & 48.08 & -0.25 & 39.44 & +0.14 \\ \midrule
20   & 44.08 & +0.11 & 22.76 & +0.43 & 32.35 & +0.45 & 44.37 & -0.45 & 49.33 & +0.72 & 34.73 & -0.41 & 48.79 & +0.46 & 39.49 & +0.19 \\ \midrule
\textbf{50}   & 43.71 & -0.26 & 23.00 & +0.67 & \textbf{34.70} & +2.80 & 44.18 & -0.64 & 49.13 & +0.52 & \textbf{37.23} & +2.09 & \textbf{48.79} & +0.46 & \textbf{40.11} & +0.81 \\ \midrule
60   & 44.02 & +0.05 & 22.74 & +0.41 & 33.82 & +1.92 & 44.29 & -0.53 & 49.36 & +0.75 & 36.69 & +1.55 & 48.54 & +0.21 & 39.92 & +0.62 \\ \midrule
80   & 43.84 & -0.13 & 22.86 & +0.53 & 33.78 & +1.88 & 44.25 & -0.57 & 49.43 & +0.82 & 36.57 & +1.43 & 48.72 & +0.39 & 39.92 & +0.62 \\ \midrule
100   & 43.75 & -0.22 & 22.87 & +0.54 & 34.00 & +2.10 & 44.17 & -0.65 & 49.36 & +0.75 & 36.60 & +1.46 & 48.64 & +0.31 & 39.91 & +0.61 \\ 
\bottomrule
\end{tabular}
}
\label{tab:ablation_loss_mad_suppl}
%\vspace{-12pt}
\end{table*}

\subsection{More Results in Ablation Studies}

\begin{table*}[t!]
\centering
\caption{Ablation study on the effect of different parameters for add $L_{cmd}$ with $L_{umd}$ on MUSES dataset~\cite{brodermann2024muses}. w/o means the framework is only trained with $L_{mad}$ + $L_{umd}$. }
\resizebox{\textwidth}{!}{
\begin{tabular}{cccccccccccccccccc}
\toprule
$\beta$ & F & \textbf{$\Delta \uparrow$} & E & \textbf{$\Delta \uparrow$} & L & \textbf{$\Delta \uparrow$} & FE & \textbf{$\Delta \uparrow$} & FL & \textbf{$\Delta \uparrow$} & EL & \textbf{$\Delta \uparrow$} & FEL & \textbf{$\Delta \uparrow$} & Mean & \textbf{$\Delta \uparrow$} \\ \midrule
w/o & 45.82 & - & 19.26 & - & 31.79 & - & 45.88 & - & 51.11 & - & 33.56 & - & 50.60 & - & 39.72 & - \\ \midrule
1 & 45.93 & +0.11 & 18.76 & -0.50 & 31.84 & +0.05 & 45.96 & +0.08 & 51.22 & +0.11 & 33.49 & -0.07 & 50.82 & +0.22 & 39.72 & 0.00 \\ \midrule
3 & 46.04 & +0.22 & 17.74 & -1.52 & 31.42 & -0.37 & 46.08 & +0.20 & 51.27 & +0.16 & 33.46 & -0.10 & 50.99 & +0.39 & 39.57 & -0.15 \\ \midrule
5 & 46.19 & +0.37 & 17.27 & -1.99 & 31.03 & -0.76 & 46.27 & +0.39 & 51.34 & +0.33 & 33.40 & -0.16 & 51.05 & +0.45 & 39.51 & -0.21 \\ \midrule
7 & 46.21 & +0.39 & 17.40 & -1.86 & 31.06 & -0.73 & 46.37 & +0.49 & 51.29 & +0.18 & 33.79 & +0.23 & 51.12 & -0.12 & 39.60 & -0.12 \\ \midrule
\textbf{10} & 46.01 & +0.19 & 19.57 & +0.31 & 32.13 & +0.34 & 46.29 & +0.41 & 51.25 & +0.14 & 35.21 & +1.65 & 51.14 & +0.54 & 40.23 & +0.51 \\ \midrule
13 & 46.57 & +0.75 & 18.24 & -1.02 & 30.88 & -0.91 & 46.74 & +0.86 & 51.09 & -0.02 & 33.88 & +0.32 & 50.76 & +0.16 & 39.74 & +0.02 \\
\midrule
15 & 46.03 & +0.21 & 14.10 & -8.90 & 31.12 & -3.58 & 45.99 & +1.81 & 50.97 & +1.84 & 31.42 & -5.81 & 50.49 & -0.11 & 38.59 & -1.13 \\
\midrule
20 & 45.95 & +0.13 & 15.19 & -7.81 & 30.61 & -4.09 & 45.80 & +1.62 & 51.19 & +2.06 & 30.55 & -6.68 & 50.41 & +1.62 & 38.53 & -1.58 \\
\bottomrule
\end{tabular}
}
\label{tab:ablation_loss_cmd_baseumd_suppl}
%\vspace{-12pt}
\end{table*}

\end{document}